\documentclass{llncs}

\usepackage{graphicx} 
\usepackage{latexsym}
\usepackage{subfig}
\usepackage{wrapfig}
\usepackage{enumitem}
\usepackage{hyperref}
\usepackage{array}
\usepackage{booktabs} 
\usepackage{color}
\usepackage{multirow}
\usepackage{booktabs}
\usepackage{amsmath}
\usepackage{amssymb} 
\usepackage{url}

\begin{document}

\title{Stance Detection in Web and Social Media: A Comparative Study}

\newcommand{\repeatthanks}{\textsuperscript{\thefootnote}}

\author{Shalmoli Ghosh\thanks{Equal contribution by authors}\inst{1}\and
        Prajwal Singhania\repeatthanks\inst{1} \and
        Siddharth Singh\repeatthanks\inst{1}  \and \\
        Koustav Rudra\thanks{The work was done when the author was a Research Associate at IIT Kharagpur}\inst{2} \and
        Saptarshi Ghosh \inst{1}
}

\institute{Department of Computer Science and Engineering,\\ Indian Institute of Technology Kharagpur, India \and
L3S Research Center, Hannover, Lower Saxony, Germany}

\authorrunning{S. Ghosh et al.}
%

\maketitle

\begin{abstract}
Online forums and social media platforms are increasingly being used to discuss topics of varying polarities where different people take different stances.  
Several methodologies for automatic stance detection from text have been proposed in literature. To our knowledge, there has not been any systematic investigation towards their reproducibility, and their comparative  performances. 
In this work, we explore the reproducibility of  several existing stance detection models, including both neural models and classical classifier-based models. 
Through experiments on two datasets -- (i)~the popular SemEval microblog dataset, and (ii)~a set of health-related online news articles -- 
we also perform a detailed comparative analysis of various methods and explore their shortcomings.
Implementations of all algorithms discussed in this paper are available at
\url{https://github.com/prajwal1210/Stance-Detection-in-Web-and-Social-Media}.
\end{abstract}


\section{Introduction}\label{sec:intro}

Online platforms such as Twitter, Facebook and discussion forums, have become popular platforms for discussing and expressing opinions about various incidents/topics. 
In this context, {\it stance} is basically an opinion expressed by an individual towards some topic or event or personality. 
For instance, in the context of a socio-political issue such as legalizing abortion in a country, some people can support the issue while some others can oppose it, and yet others can be neutral.
In today's Web, automatically identifying stance of a person from an online post authored by the person, is an important problem (which is called {\it stance detection}).
Automatic stance detection has several applications~\cite{mohammad2016semeval}, including understanding the public opinion towards a specific socio-political issue, understanding the credibility of an online post based on whether it is supported by authentic users, and so on.

According to earlier works, stance detection can be of two types -- (i)~Multi-target Stance Detection, and (ii)~Target (single) specific Stance Detection. 
Multi-target stance detection aims at jointly detecting stances towards multiple related targets. This problem was introduced by Sobhani {\it et al.}~\cite{sobhani2017dataset} and has been studied by many later works~\cite{liu2018two,sobhani2019exploring,wei2018multi}. 
In target-specific stance detection, the targets are considered separately and individually.  
In this work, we will be focusing on the problem of target-specific stance detection.
Many algorithms have been proposed for  target-specific stance detection; see Section~\ref{sec:related} for a survey on such methods. 
However, to our knowledge, there has neither been any systematic comparison of these methods, nor any investigation of how reproducible these methods actually are. 
The present paper attempts to bridge this gap. 

In this work, we explore {\it seven} target-specific stance detection models, out of which we implemented six (and publicly available code was used for the other model).
We first investigate the reproducibility of the models. We then apply them on two datasets -- (1)~the standard SemEval microblog dataset, and (2)~a formal text dataset of health-related articles. 
We also develop a new method that applies the recently developed BERT model~\cite{devlin2018bert} for stance detection, and compare its performance to that of the existing methods. 
Implementations of all algorithms discussed in this paper are available at:\\
\url{https://github.com/prajwal1210/Stance-Detection-in-Web-and-Social-Media}.



\section{Related work}\label{sec:related}

In recent times, various works have tackled stance classification in different fields such as 
controversy detection~\cite{jang2016improving}, news articles~\cite{ferreira2016emergent}, 
student essays~\cite{faulkner2014automated}, 
and so on. A reecent work~\cite{lai2018stance} also studied stance detection from a diachronic perspective.
The earlier models used traditional feature engineering-based methods, while the more recent models use deep neural architectures. We survey some stance detection models in this section.

\vspace{2mm}
\noindent {\bf Stance detection using traditional feature engineering:}
Various works on stance detection use traditional feature engineering. For instance, 
Sen {\it et al.}~\cite{sen2018stance} proposed a novel set of features with SVM model and a feedforward neural network model. 
HaCohen-kerner {\it et al.}~\cite{hacohen2017stance} used 18 features including character skip-ngrams and character ngrams. 
Dilek {\it et al.}~\cite{kuccuk2018stance} used unigrams, bigrams, hashtags, external links, emoticons, and named entities as features to a SVM model. 
Dey {\it et al}~\cite{dey2017twitter} proposed a two-phase SVM architecture with borrowed and novel feature sets.

It can be seen that most of the feature engineering based methods -- including the baseline methods given by SemEval challenge~\cite{mohammad2016semeval} (that standardized the problem of stance detection over microblogs) -- use SVM as a classifier. 
Hence, for our comparative analysis, we have chosen two SVM-based models.

\vspace{2mm}
\noindent {\bf Stance detection using neural models:}
In recent years, there have been many works using neural models for stance detection. Du {\it et al.}~\cite{du2017stance} used an attention based model for stance classification.
Mitre {\it et al.}~\cite{zarrella2016mitre}, the winning team of the SemEval 2016 Task 6A challenge, proposed a transfer learning method with features learned via distant supervision on two large unlabelled datasets.
Wei {\it et. al.}~\cite{Wei2016pkudblabAS}), the second position holders of SemEval 2016 Task 6A challenge, used Kim's CNN. 
Chen {\it et al.}~\cite{chen2016utcnn} applied neural network model to classify stance of social media posts by considering users' taste, topics' taste and user comments on posts, 
whereas Dey {\it et al.}~\cite{dey2018topical} used a two phase LSTM model. 

Neural models for stance detection can be divided into a few informative categories like attention-based~\cite{du2017stance}, convolution-based~\cite{Wei2016pkudblabAS} and word embedding-based models~\cite{hochreiter1997long}. 
For our comparative analysis, we have chosen one representative from each category, along with a recent pre-trained model named BERT~\cite{devlin2018bert}.

\vspace{2mm}
\noindent {\bf Surveys on stance detection models:}
There have been some relevant surveys as well. 
Zubiaga {\it et al.}~\cite{zubiaga2018detection} discussed various stance detection approaches for rumour detection and resolution. Several stance detection approaches were compared on Spanish and Catalan datasets in the StanceCat task~\cite{taule2017overview}. 
Wang {\it et al.}~\cite{wang2019survey} analysed the shortcomings of different stance detection models. But this survey did {\it not} actually compare performances of different methods over specific datasets, and also did not explore the reproducibility of different models, as is done in the present paper.


\section{Dataset and Preprocessing}
\label{sec:dataset}

In this section, we describe the datasets used for the comparative analysis, and the preprocessing used over the datasets. 

\subsection{Datasets}

We have used two types of publicly available stance-detection datasets: 

\vspace{2mm}
\noindent \textbf{(1) SemEval 2016 Task 6A Dataset~\cite{mohammad2016semeval}:}
contains microblogs (tweets) data related to the following 5 topics --
(i)~Atheism (AT),
(ii)~Climate Change is a real concern (CC),
(iii)~Feminist Movement (FM),
(iv)~Hillary Clinton (HC), and
(v)~Legalization of Abortion (LA).
For each topic, we have used the official train-test split, as used in the SemEval 2016 challenge. 

\vspace{2mm}
\noindent \textbf{(2) Multi Perspective Consumer Health Query (MPCHI) Data:} This dataset, taken from~\cite{sen2018stance}, comprises of formal texts (sentences collected from top-ranked articles corresponding to queries issued on a specific Web search engine) corresponding to the following 5 queries (claims) --
(i)~MMR vaccination can cause autism (MMR),
(ii)~E-cigarettes are safer than normal cigarettes (EC),
(iii)~Women should take HRT post menopause (HRT),
(iv)~Vitamin C prevents common cold (VC),
(v)~Sun exposure leads to skin cancer (SC).
We split each dataset of MPCHI into train and test sets in the same proportion as in the SemEval data. 

\vspace{2mm}
\noindent Each dataset contains texts annotated with one of three classes -- Favor (supports the topic/claim), Against (opposes the topic/claim), and None (neutral to the topic/claim). 
Table~\ref{table:dataset} reports the statistics of all 10 datasets, and Table~\ref{tab:text} shows some example posts from the datasets.

\begin{table}[tb]
\footnotesize
\center

\resizebox{!}{60pt}{
\begin{tabular}{|c||c||c|c|c||c|c|c|}
\hline
\textbf{Dataset} & \textbf{Topic} & \multicolumn{3}{c||}{\textbf{\# Training Instances}}  & \multicolumn{3}{c|}{\textbf{\# Test Instances}}\\
\cline{3-8}
 & &  FAVOR & AGAINST & NONE &  FAVOR & AGAINST & NONE \\
\hline
 & AT & 92 & 304 & 117 & 32 & 160 & 28 \\ \cline{2-8}
 & CC & 212 & 15 & 168 & 123 & 11 & 35 \\ \cline{2-8}
SemEval & FM & 210 & 328 & 126 & 58 & 183 & 44 \\ \cline{2-8}
& HC & 112 & 361 & 166 & 45 & 172 & 78 \\ \cline{2-8}
& LA & 105 & 334 & 164 & 46 & 189 & 45 \\
\hline
\hline
& MMR & 48 & 61 & 72 & 24 & 33 & 21 \\ \cline{2-8}
& SC & 68 & 51 & 117 & 35 & 26 & 42 \\ \cline{2-8}
MPCHI & EC & 60 & 118 & 111 & 33 & 47 & 44 \\ \cline{2-8}
& VC & 74 & 52 & 68 & 37 & 16 & 31 \\ \cline{2-8}
& HRT & 33 & 95 & 44 & 9 & 41 & 24 \\
\hline
\end{tabular}}
\caption{\textbf{Statistics of the datasets (divided into training and test sets).}}
\label{table:dataset}
\vspace{-3mm}
\end{table}

\begin{table*}[tb]
\footnotesize
\centering

    \begin{tabular}{|p{0.85\textwidth}|c|} \hline
        \textbf{Tweet / Text} & \textbf{Label}\\ \hline
    \multicolumn{2}{|c|}{{\bf Tweets from SemEval Dataset AT (Atheism)}} \\ \hline
        
        All that is needed for God for something to happen is to say "\#Be" and it is; for God is capable of all things. \#God created \#trinity \#SemST & AGAINST\\ \hline
        
       Absolutely fucking sick \& tired of the religious and their "We're persecuted" bollocks\! So f**king what? Pissoff! \#SemST & FAVOR \\ \hline
       
    In other related news. Boko Haram has killed over 200 people in the last 48hrs. \#SemST & NONE  \\ \hline
    


    \multicolumn{2}{|c|}{{\bf Texts from MPCHI Dataset HRT}} \\ \hline
    A 2002 study called the Women's Health Initiative (WHI), designed to explore the benefits and risks of combined estrogen-progestin HRT was halted three and a half years before the intended end of the study period, because researchers observed a 26 percent increase in the relative risk of breast cancer. & AGAINST\\ \hline
    
    HRT can also help to lower the risk of osteoporosis  and prevent some of the long term health problems associated with early menopause. & FAVOR\\ \hline
    
    Ovarian cancer is the fifth most common cause of cancer death among women in the UK, accounting for six per cent of all female deaths from cancer. & NONE \\ \hline
    
    
    \end{tabular}
    \caption{\textbf{Examples of posts from some of the datasets}}
    \label{tab:text}
    \vspace{-5mm}
\end{table*}

\subsection{Preprocessing Methodology}

Different prior works have used different preprocessing methods. To ensure a fair comparison among different models, we apply the same preprocessing before applying the models.

\vspace{1mm}
\noindent\textbf{Standard preprocessing:} We perform standard preprocessing steps such as case-folding, stemming (using Porter stemmer), and stopword removal. 
However, note that stemming and stop-word removal are {\it not} performed while using neural models that rely on pre-trained embeddings (since the stemmed versions of terms would not probably be found in the pre-trained embeddings).

\vspace{1mm}
\noindent\textbf{Exclusive preprocessing for microblogs:} We perform the following preprocessing only for the microblog datasets (across all stance detection models):

\noindent \textbf{(1) Normalization:} We normalize the text using the method proposed by Han {\it et. al}~\cite{han2011lexical}. This helps us to deal with abbreviations and out-of-vocabulary words. For example, the term `{\it aaf}' is expanded as `{\it as a friend}'.

\noindent \textbf{(2) Hashtag Preprocessing:} Users primarily use hashtags in tweets to convey their sentiments~\cite{han2011lexical}. Hashtags are often created by concatenating several individual words. For example, \#powertowomen is a popular hashtag used during the `Feminist Movement'. Such hashtags are usually marked as OOV (Out Of Vocabulary) words by standard NLP tools. In this paper, we have used Wordninja package\footnote{\url{https://github.com/keredson/wordninja}} to split such combined texts into most probable constituent word sequences. For example, `\textbf{\#powertowomen}' may be splitted as  `\textbf{\# power to women}' or `\textbf{\# power tow omen}'. However, the algorithm returns the first one because it is more probable than the later one.

Incorporating the tweet normalization and hashtag preprocessing steps in this work have resulted in improved performance of existing models. Later in Section~\ref{sec:result}, we have shown the effect of this preprocessing.


\section{Reproducing a Selection of Stance Detection Methods}\label{sec:method}

For our present study, we selected a few representative methods from the two groups of methods stated in Section~\ref{sec:related}.
In this section, we describe the challenges in reproducing the methods and possible ways to overcome the challenges.
Note that, we used codes provided by the authors for the first model, while all the other models were reproduced by us.

\vspace{2mm}
\noindent\textbf{(1) Convolutional Neural Networks:}
This method~\cite{Wei2016pkudblabAS}, which uses Kim's 1-D CNN-based sentence classification model~\cite{DBLP:journals/corr/Kim14f}, performed second-best in the SemEval stance detection task (Task 6A). We used the code that has been made available by the authors.\footnote{\url{https://github.com/nestle1993/SE16-Task6-Stance-Detection}}
Note that, we applied our pre-processing techniques on the dataset before applying this model, and this step significantly improves the performance (see Section~\ref{sec:result}).

\vspace{2mm}
\noindent\textbf{(2) Target-Specific Attention Neural Network [TAN]: } Du {\it et al.}~\cite{du2017stance} proposed a novel bidirectional LSTM-based attention mechanism. 
We briefly describe the architecture below. 
A target sequence of length $N$ is represented as $[z_1, z_2, \ldots , z_N]$
where $z_n \epsilon R^{d^{'}}$ is the $d^{'}$-dimensional vector of the $n$-th word in the target sequence. 
The target-augmented embedding of a word $t$
for a specific target $z$ is $e^{z}_{t}=x_{t} \odot z$ where $\odot$ is the vector concatenation operation. 
The dimension of $e^z_t$ is $(d+d^{'})$. An affine transformation maps the $(d+d^{'}$)-dimensional target-augmented embedding of each word to a scalar value as per Eqn.~\ref{eqn:eqn13}: 
\begin{equation}
a^{'}_{t} = W_{a}e_{t}^{z} + b_{a}   
\label{eqn:eqn13}
\end{equation} 
where $W_a$ and $b_a$ are the parameters of the bypass neural network. The attention vector $[a^{'}_{1}, a^{'}_{2}, \ldots, a^{'}_{T}]$ undergoes a softmax transformation to get the final attention signal vector (Eqn.~\ref{eqn:eqn14}):
\begin{equation}
    a_{t} = softmax(a_{t}) = \frac{e^{a^{'}_{t}}}{\sum_{i=1}^{T}e^{a^{'}_{i}}}
    \label{eqn:eqn14}
\end{equation}

\vspace{1mm}
\noindent {\bf Challenges in Reproducibility:}
Du {\it et al.}~\cite{du2017stance} mentioned that they trained embeddings on a manually scraped corpus, but they neither released the corpus nor the embeddings. We have used the pre-trained Glove 6B 300d embeddings for this purpose.\footnote{\url{https://nlp.stanford.edu/projects/glove/}}
Additionally, whether dropout is used and what activation function is used in the middle layers were not mentioned in~\cite{du2017stance}. We have used dropout and ReLU activation function.

\vspace{1mm}
\noindent {\bf An observation about the TAN model:}
Du. {\it et al}~\cite{du2017stance} claim that using the target-augmented embeddings enable the model to make \textit{``full use of the target information in stance detection''} (quoted from~\cite{du2017stance}). 
However, we believe that this architecture does {\it not} take advantage of the target information at all. We give a simple proof for our claim:

\begin{theorem}
The bypass neural network in the TAN is unaffected by the target information, i.e., $\frac{da_{t}}{dz} = 0$ .
\end{theorem}

\noindent {\bf Proof:}
From Eqn.~\eqref{eqn:eqn13}, we have:
    $a^{'}_{t} = W_{a}e_{t}^{z} + b_{a} \implies
    a^{'}_{t} = W_{a}(x_{t} \odot z) + b_{a}$
\[
\because W_{a} = W_{ax} \odot W_{az}\: (\mbox{where} \: W_{ax} \epsilon R^{d} \: \mbox{and} \: W_{az} \epsilon R^{d^{'}}) :
\]
\[
\therefore  a^{'}_{t} = W_{ax}\cdot x_{t} + W_{az}\cdot z + b_{a}
 \]   
Now, from Eqn.~\eqref{eqn:eqn14}, we have:
\[
    a_{t} = \frac{e^{a^{'}_{t}}}{\sum_{i=1}^{T}e^{a^{'}_{i}}} =   \frac{e^{W_{ax}\cdot x_{t} + W_{az}\cdot z + b_{a}}}{\sum_{i=1}^{T}e^{W_{ax}\cdot x_{i} + W_{az}\cdot z + b_{a}}}  = \frac{e^{W_{ax}\cdot x_{t} }}{\sum_{i=1}^{T}e^{W_{ax}\cdot x_{i} }} 
\]
\[
\therefore \frac{da_{t}}{d{z}} = 0
\]
We also back our claim with an empirical experiment, wherein we do {\it not} augment the target embeddings to the word-embeddings in the bypass neural network, i.e., we use $e^{z}_{t}=x_{t}$ (instead of $e^{z}_{t}=x_{t} \odot z$ in the original model). 
We call this architecture the \textbf{TAN-}. 
We show later in the paper that results obtained by both the TAN and TAN- architectures are very similar.

\vspace{2mm}
\noindent\textbf{(3) Recurrent Neural Network with Long Short Term Memory(LSTM):}
In the earlier TAN paper~\cite{du2017stance}, one of the baselines was LSTM without target-specific embedding and target-specific attention. In this work we have also reproduced this LSTM-based method.

\vspace{1mm}
\noindent\textbf{Challenges in Reproducibility:}
The challenges faced were same as TAN model and we took the same steps as for the TAN model~\cite{du2017stance}.

\vspace{2mm}
\noindent\textbf{(4) SVM-based SEN Model: }
Sen {\it et al.}~\cite{sen2018stance} proposed a SVM based stance detection model using five sets of features -- stance vector, textual entailment, sentiment, medical knowledge based feature and a standard context based BoW feature. 
The stance vector is created on a sentence level based on an assumption~\cite{agrawal2012unsupervised} that the main information present in a sentence revolves around some particular parts-of-speech like the Nouns, Adjectives, Verbs, Adverbs. Thus these parts-of-speech are the main building blocks of the stance expressed by a sentence towards a particular claim.
To identify the sentiment feature (positive or negative or neutral), we used a standard sentiment analyzer given in Stanford CoreNLP Toolkit to obtain the sentiment for a sentence.

Note that, we made one change while implementing this model. For the textual entailment feature, the original paper~\cite{sen2018stance} used the Excitement Open Platform (EOP)~\cite{pado2015design}.
We initially tried using EOP, but later we observed that results improve if textual entailment is estimated with Tensor Flow\footnote{\url{https://tinyurl.com/entailment-with-tensorflow}}
where textual entailment is estimated using word vectorization, recurrent neural networks with LSTM  and dropout as a regularization method.

Finally, the medical knowledge-based features were extracted using a tool called SemRep (\url{https://semrep.nlm.nih.gov}) along with the help of a medical knowledge-base called UMLS~\cite{bodenreider2004unified}. 
This feature is specifically for use with the MPCHI datasets (on which Sen {\it et al.}~\cite{sen2018stance} performed their experiments). The medical feature is not used for SEMEVAL dataset as it is not related to health informatics.

\noindent\textbf{Challenges in Reproducibility:} In the BoW feature, we used word-unigrams, since the exact value of $n$ for n-grams was not mentioned in the original paper~\cite{sen2018stance}.

\vspace{2mm}
\noindent\textbf{(5) Two-step SVM:}
 Dey {\it et al.}~\cite{dey2017twitter} proposed a two-step stance detection approach. In the first step, they find whether a tweet is relevant to the given claim, and in the next step they detect the stance (if the tweet is relevant). 
The first step uses features such as Weighted MPQA Subjectivity-Polarity Classification and Wordnet Based Potential Adjective Recognition, whereas the second phase comprises of Sentiwordnet and MPQA Based Sentiment Classification, Frame Semantics, Target Detection, Word n-Grams and Character n-Grams. 

\vspace{1mm}
\noindent {\bf Challenges in Reproducibility:} 
According to Dey {\it et al.}~\cite{dey2017twitter}, the two most important features are (i)~Wordnet Based Potential Adjective Recognition and (ii)~Frame Semantics. Especially, in the second phase, frame semantics is the most decisive feature. 
This feature attempts to estimate the relative importance of multiple clauses present in a tweet, where the clauses are considered to be separated by `connector words'. 
However, there is lack of clarity about this feature. It is written that ``We assign more weightage to the more important clause, in case connector words are present in the sentence.'' (quoted from~\cite{dey2017twitter}). 
But it is not clarified how exactly the relative weights of the clauses are decided. 
Due to this lack of clarity, we could not implement the frame semantics feature of this model. We have implemented all other features except frame semantics.

\vspace{1mm}
\noindent {\bf A general challenge in reproducibility across all models:}
Almost none of the prior works described in this section stated the exact values of the hyperparameters in the models. Hence, we adopted the same approach as Du {\it et al.}~\cite{du2017stance} --  hyperparameters were tuned using 5-fold cross-validation on the training set (of each dataset).


\section{Using BERT for Stance Detection}
\label{sec:bert}

Apart from experimenting with existing stance detection methods, we have applied a recently developed deep learning model named BERT~\cite{devlin2018bert} (Bidirectional Encoder Representations from Transformers - developed by Google AI Language group) for stance detection.
To the best of our knowledge, no prior work has applied BERT for stance detection.
 
BERT models pre-trained on large unlabeled corpora using bidirectional language modelling have been released by Google. 
This training is made possible by masking $15\%$ of the input words, and using the corresponding final layer hidden states to predict these words.
The pre-trained BERT model can be fine-tuned with just one additional output layer to create state-of-the-art models for a wide range of tasks, such as question answering and language inference, without substantial task-specific architecture modifications.

In this work, we used a pre-trained BERT (Large-Uncased) model. The input text is fed to the BERT model which generates representations of the words in the text through multiple transformer layers.  We have then fed the output of the first head of the final layer of BERT through a randomly initialized feed-forward layer along with softmax and fine-tuning the network on the task-specific data.


\section{Experimental Setup}
\label{sec:exp}

In this section we briefly describe the experimental settings, including the baselines, evaluation metrics, and the method used for parameter tuning.

\noindent{\textbf{Hyper-parameter Tuning:}}
As stated in Section~\ref{sec:method}, almost none of the prior works specify all the hyperparameter values.
We tuned all hyperparamters (that are not stated in the respective papers) using 5-fold cross-validation on the training set. In case a hyperparameter value is specified in a paper, the said value is used. Model-wise hyperparameter values that were tuned by us are mentioned in Table~\ref{tab:hyper} (using same notations as in our codes (for those not mentioned in the respective papers)).

\vspace{2mm}
\noindent{\textbf{Vote Scheme:}}
We use the vote scheme proposed in~\cite{Wei2016pkudblabAS} for prediction on the test set. For each model, we run ten parallel epochs, whose validation sets are randomly selected from the training
set and are non-overlapping. 
According to~\cite{Wei2016pkudblabAS}, in each epoch, some iterations are deliberately chosen to predict the test set. Then, when this epoch ends, for every sentence in the test set, the label which appears most frequently in these predictions as the result of this epoch is appointed. Finally, when ten epochs end, voting happens within results of these ten epochs by the same method described above to determine the final labels.
Performing multiple times independently and voting twice provides a robust mechanism for predicting. 
Note that the voting scheme is used for TAN, TAN-, CNN and LSTM models only.

\begin{table*}[tb]
\footnotesize
    \centering
    \begin{tabular}{|p{0.2\textwidth}|p{0.8\textwidth}|} \hline
        \textbf{Model} & \textbf{Hyperparameters}  \\ \hline
        TAN and TAN- & Learning\_rate: 5e-4, batch\_size:50, dropout: 0.5, L2:[(AT, HRT):1.25, (CC, LA, HC):1, FM:0.75, (MMR,  SC, VC, EC):0.25], epochs:[(AT, LA, VC):40-50,(CC, FM,  HC, MMR, HRT, SC, EC):50-60]\\ \hline
       
        LSTM & Learning\_rate: 5e-4, batch\_size:50, dropout: 0.5, L2:[(AT,  HC, VC, EC):0.25, (CC, LA, MMR, HRT, SC):0.5, FM:0.75], epochs:[(AT, CC, FM, LA, HC, SC):50-60, (MMR, HRT, VC, EC):30-40]\\ \hline
        CNN & Dropout: 0.5, Learning\_rate\_decay : 0.95 , Squared norm limit:[(AT, FM, LA, MMR, VC, EC):7, (CC, HC, HRT):8, SC:9] \\ \hline
        BERT & Learning\_Rate:2e-5, Num\_Train\_Epochs:50, Warmup\_Proportion:0.1, Max\_Seq\_Length:128\\ \hline
        SEN & gamma:0.001(for both MPCHI and SemEval), rest as in paper ~\cite{sen2018stance}\\ \hline
        Two-step SVM & As given in paper ~\cite{dey2017twitter}\\ \hline
         
    \end{tabular}
    \caption{\textbf{Hyperparameters of the models. For some hyperparameters, the values are different for different datasets.}}
    \label{tab:hyper}
\end{table*}

\vspace{1mm}
\noindent \textbf{Performance Metric:} To evaluate the performance of all models, we use the same metric as used by the official SemEval 2016 Task A~\cite{mohammad2016semeval} -- the macro-average of the F1-score for `favor' and `against' classes.


\section{Results and Analysis}
\label{sec:result}

This section describes the comparative analyses of the different stance detection models, and also reports some error analysis. 

\vspace{2mm}
\noindent \textbf{Effect of Preprocessing on existing methods:}
In this work, we applied some tweet-specific preprocessing (tweet normalization and hashtag preprocessing) on the SemEval datasets (as stated in Section~\ref{sec:dataset}). 
Table~\ref{tab:preprocessing} reports the performance of some of the models, as reported in the original paper (that proposed a model) and after this tweet-specific preprocessing.
We see that the performance of the existing methods improves significantly due to this preprocessing.


\begin{table}[tb]
    \centering
    \scalebox{0.95}{
    \begin{tabular}{|c||c|c|}  \hline
     \textbf{Method} & \textbf{Metric value reported} & \textbf{Metric value with} \\ 
     & \textbf{by original paper} & \textbf{our preprocessing}\\ \hline
        
        TAN  & 0.6879 ~\cite{du2017stance} & \textbf{\textcolor{blue}{0.690}}\\ \hline
        
        LSTM  &  0.6321 ~\cite{du2017stance} & \textbf{\textcolor{blue}{0.687}}\\ \hline
         
        CNN & 0.6733 ~\cite{Wei2016pkudblabAS} & \textbf{\textcolor{blue}{0.706}}\\ \hline
    \end{tabular}}
    \caption{\textbf{Results showing the effect of our preprocessing on previous models (computed over all the datasets of SEMEVAL)}}
    \label{tab:preprocessing}
\end{table}

\vspace{2mm}
\noindent \textbf{Comparative analysis:}
Table~\ref{table:twitter_result} and Table~\ref{table:MPCHI_result} describe the performances of all models on SemEval dataset and MPCHI dataset respectively. 
Since we could not reproduce the Frame Semantics feature of the two-step SVM model~\cite{dey2017twitter}, we have reported both the performances of our implementation and that reported in the original paper~\cite{dey2017twitter} for the SemEval datasets (the original paper worked only on the SemEval datasets, not the MPCHI datasets).

It is clearly seen that the overall metric of BERT model is far better than that of other competing models. Apart from the BERT model, all other models perform better in case of SemEval dataset consistently. 
This is possibly because the size of MPCHI dataset is much smaller than that of SemEval dataset, and thus neural models might not train well over such small datasets. 
Also we observed that the CNN model performs well for shorter tweets (of length 5-10 words) while BERT works well for longer ones, since BERT is developed to capture context information over longer texts (details omitted due to lack of space). 

Note that the results of TAN and TAN- models are very comparable as claimed in Section~\ref{sec:method}; in fact, as per the overall metric, the TAN- model performs slightly better than TAN for both types of datasets.

\begin{table*}[tb]
\small
\centering
\footnotesize
\scalebox{0.85}{
\begin{tabular}{|p{0.4\textwidth}||c|c|c|c|c|c|c|}
\hline
 \textbf{Model}&\textbf{AT} & \textbf{CC} & \textbf{LA} & \textbf{FM} & \textbf{HC}  &\textbf{TOTAL} \\ \hline

TAN ~\cite{du2017stance} &  0.628 & 0.430 & 0.567 & 0.590 & \textbf{\textcolor{blue}{0.728}} & 0.690\\ \hline
TAN- &  0.638 & 0.440 & 0.572 & 0.542 & 0.724 & 0.692\\ \hline
LSTM~\cite{du2017stance} &  0.629 & 0.429 & 0.628 & 0.571 & 0.611 & 0.687\\ \hline
SEN~\cite{sen2018stance} & 0.590 & 0.39 & 0.575 & 0.510 & 0.565 & 0.630 \\ \hline
CNN~\cite{Wei2016pkudblabAS} &  0.641 & 0.445 & \textbf{\textcolor{blue}{0.684}} & 0.552 & 0.675 & 0.706\\ \hline
BERT~\cite{devlin2018bert} &  \textbf{\textcolor{blue}{0.743}}& \textbf{\textcolor{blue}{0.446}} & 0.657 & \textbf{\textcolor{blue}{0.650}} & 0.713 & \textbf{\textcolor{blue}{0.751}}\\ \hline
Two-step SVM (without Frame Semantics) & 0.410 & 0.419 & 0.436 & 0.496 & 0.488 & 0.631\\ \hline
Two-step SVM (as reported in~\cite{dey2017twitter}) & 0.725 & 0.535 & 0.836 & 0.787 & 0.797 & 0.744\\ \hline
\end{tabular}}
\caption{{\bf Stance classification results on SEMEVAL datasets. Highest values marked in blue and boldface.}}
\label{table:twitter_result}
\end{table*}

\begin{table*}[tb]
\small
\centering
\scalebox{0.85}{
\begin{tabular}{|p{0.3\textwidth}||c|c|c|c|c|c|c|}
\hline
 \textbf{Model} & \textbf{HRT} & \textbf{EC} &
      \textbf{VC} & \textbf{SC} &\textbf{MMR}  &\textbf{TOTAL} \\
      \hline

TAN ~\cite{du2017stance} &  0.347 & 0.580 & 0.421 & 0.507 & 0.671 & 0.586\\ \hline
TAN- &  0.569 & 0.583 & 0.578 & 0.468 & 0.608 & 0.589\\ \hline
LSTM ~\cite{du2017stance} &  0.464 & 0.609 & 0.592 & 0.575 & 0.665 & 0.631\\ \hline
SEN~\cite{sen2018stance} & 0.480 & 0.605 & 0.405 & 0.445 & 0.615 & 0.540 \\ \hline
CNN ~\cite{Wei2016pkudblabAS}&  0.359 & 0.539 & 0.524 & 0.252 & 0.524 & 0.551\\ \hline
BERT~\cite{devlin2018bert} & \textbf{\textcolor{blue}{0.669}}  & \textbf{\textcolor{blue}{0.780}} & \textbf{\textcolor{blue}{0.647}} & \textbf{\textcolor{blue}{0.769}} & \textbf{\textcolor{blue}{0.782}} & \textbf{\textcolor{blue}{0.756}}\\ \hline
Two-step SVM (without Frame Semantics) & 0.470 & 0.297 & 0.409 & 0.293 & 0.455 & 0.519\\ \hline

\end{tabular}}
\caption{{\bf Stance classification results on MPCHI datasets. Highest values marked in blue and boldface.}}
\label{table:MPCHI_result}
\end{table*}

\vspace{2mm}
\noindent \textbf{Where all models fail:}
We have considered the labels predicted by all the seven models, and checked those tweets/texts where all the models fail (i.e., no model was able to give the correct label).
In total, there are 72 tweets (across all SemEval datasets) and 42 posts (across all MPCHI datasets) where all models classified wrongly;
Table~\ref{tab:all_miss} shows some examples of such tweets/text.
We manually observed these misclassified tweets and text, and made the following observations.
\begin{itemize}
    \item In case of tweets (the SemEval datasets) the errors were mostly on tweets that contain (i)~sarcastic comments~\cite{sulis2016figurative}, and (ii)~questions. 
    \item In case of the MPCHI dataset, there were some posts containing health-related facts, which actually express no stance w.r.t the target. All the models were unable to capture this notion. 
    It is possible that the stance can be understood by a human having a lot of contextual background knowledge; however, it is difficult to understand the stance just from what is mentioned in the tweet/text.
\end{itemize}

\begin{table*}[tb]
\footnotesize
    \centering
    \scalebox{0.95}{
    \begin{tabular}{|p{0.13\textwidth}|c|p{0.8\textwidth}|}
    \hline
    \textbf{Reason} & \textbf{Dataset} & \textbf{Tweet/Text} \\ \hline
  Sarcasm & FM & I like girls. They just need to know there place. \#SemST \\ \cline{2-3}
  & CC & @JustinTrudeau Hey Justin I will give you 50 cents if you stop talking about climate `Change' \#Ottawa \#davidsuzuki \#cbc \#SemST \\ 
  \hline
  
  & HC & Do you Progressives know how dangerously close you are to suppressing free speech? Stop it.   \#inners  \#readyforhillary \#SemST \\ \cline{2-3}
 Question & FM & @BOZARbrussels is this how \@UN\_Women sees \#genderequality ? Only \#women with arms like \#men ?\#stopmarriagebill \#fakecases \@UN \#SemST\\ \cline{2-3}
 \hline

 Insufficient & EC & E-Cigarettes contain ONLY nicotine.\\ \cline{2-3}
 signal to & HRT & There are also ne therapies, such as progestogens and testosterone.\\ \cline{2-3}
 target & EC & Public health officials counter that it's too early to know very much about the health effects of e-cigarettes, especially on young people.\\ \hline
 
 
    \end{tabular}}
    \caption{{\bf Examples of tweets/texts misclassified by all models}}
    \label{tab:all_miss}
\end{table*}


\section{Conclusion} \label{sec:conclu}

To our knowledge, this is the first analysis of reproducibility of existing stance detection methods, as well as a systematic comparison of stance detection methods over two different type of datasets. 
We observed that the BERT pre-trained model can perform better stance detection than many existing methods.
We see that no single method is able to give very high metric value  over all datasets; this observation motivates us to explore  
some combination of methods (ensemble methods) for stance detection in future.

\vspace{5mm}
\noindent {\bf Acknowledgement:} The work is partially supported by a project titled ``Building Healthcare Informatics Systems Utilising Web Data'' funded by Department of Science \& Technology, Government of India.


\begin{thebibliography}{10}
\providecommand{\url}[1]{\texttt{#1}}
\providecommand{\urlprefix}{URL }
\providecommand{\doi}[1]{https://doi.org/#1}

\bibitem{agrawal2012unsupervised}
Agrawal, A., An, A.: Unsupervised emotion detection from text using semantic
  and syntactic relations. In: Proc. IEEE/WIC/ACM Joint Conferences on Web
  Intelligence and Intelligent Agent Technology (2012)

\bibitem{bodenreider2004unified}
Bodenreider, O.: The unified medical language system (umls): integrating
  biomedical terminology. Nucleic acids research  \textbf{32}(suppl\_1),
  D267--D270 (2004)

\bibitem{chen2016utcnn}
Chen, W.F., Ku, L.W.: Utcnn: a deep learning model of stance classification on
  social media text. arXiv preprint arXiv:1611.03599  (2016)

\bibitem{devlin2018bert}
Devlin, J., Chang, M.W., Lee, K., Toutanova, K.: Bert: Pre-training of deep
  bidirectional transformers for language understanding. arXiv:1810.04805
  preprint  (2018)

\bibitem{dey2017twitter}
Dey, K., Shrivastava, R., Kaushik, S.: {Twitter Stance Detection -- A
  Subjectivity and Sentiment Polarity Inspired Two-Phase Approach}. In: {Proc.
  IEEE Conference on Data Mining Workshops (ICDMW)} (2017)

\bibitem{dey2018topical}
Dey, K., Shrivastava, R., Kaushik, S.: {Topical Stance Detection for Twitter: A
  Two-Phase LSTM Model Using Attention}. In: Proc. European Conference on
  Information Retrieval (2018)

\bibitem{du2017stance}
Du, J., Xu, R., He, Y., Gui, L.: Stance classification with target-specific
  neural attention networks. {Proc. International Joint Conferences on
  Artificial Intelligence (IJCAI)} (2017)

\bibitem{faulkner2014automated}
Faulkner, A.: Automated classification of stance in student essays: An approach
  using stance target information and the wikipedia link-based measure. Science
   \textbf{376}(12), ~86 (2014)

\bibitem{ferreira2016emergent}
Ferreira, W., Vlachos, A.: Emergent: a novel data-set for stance
  classification. In: {}Proc. NAACL-HLT

\bibitem{hacohen2017stance}
HaCohen-Kerner, Y., Ido, Z., Ya’akobov, R.: Stance classification of tweets
  using skip char ngrams. In: Joint European Conference on Machine Learning and
  Knowledge Discovery in Databases. pp. 266--278. Springer (2017)

\bibitem{han2011lexical}
Han, B., Baldwin, T.: Lexical normalisation of short text messages: Makn sens
  a\# twitter. In: Proc. Annual Meeting of the Association for Computational
  Linguistics: Human Language Technologies (2011)

\bibitem{hochreiter1997long}
Hochreiter, S., Schmidhuber, J.: Long short-term memory. Neural computation
  \textbf{9}(8),  1735--1780 (1997)

\bibitem{jang2016improving}
Jang, M., Allan, J.: Improving automated controversy detection on the web. In:
  {Proc. ACM SIGIR conference} (2016)

\bibitem{DBLP:journals/corr/Kim14f}
Kim, Y.: Convolutional neural networks for sentence classification. CoRR
  \textbf{abs/1408.5882} (2014), \url{http://arxiv.org/abs/1408.5882}

\bibitem{kuccuk2018stance}
K{\"u}{\c{c}}{\"u}k, D., Can, F.: Stance detection on tweets: An svm-based
  approach. arXiv preprint arXiv:1803.08910  (2018)

\bibitem{lai2018stance}
Lai, M., Patti, V., Ruffo, G., Rosso, P.: Stance evolution and twitter
  interactions in an italian political debate. In: International Conference on
  Applications of Natural Language to Information Systems. pp. 15--27. Springer
  (2018)

\bibitem{liu2018two}
Liu, H., Li, S., Zhou, G.: Two-target stance detection with target-related zone
  modeling. In: Proc. China Conference on Information Retrieval. Springer
  (2018)

\bibitem{mohammad2016semeval}
Mohammad, S., Kiritchenko, S., Sobhani, P., Zhu, X., Cherry, C.: Semeval-2016
  task 6: Detecting stance in tweets. In: Proc. International Workshop on
  Semantic Evaluation (SemEval-2016). pp. 31--41 (2016)

\bibitem{pado2015design}
Pad{\'o}, S., Noh, T.G., Stern, A., Wang, R., Zanoli, R.: Design and
  realization of a modular architecture for textual entailment. Natural
  Language Engineering  \textbf{21}(2),  167--200 (2015)

\bibitem{sen2018stance}
Sen, A., Sinha, M., Mannarswamy, S., Roy, S.: Stance classification of
  multi-perspective consumer health information. In: Proc. ACM India Joint
  Conference on Data Science and Management of Data (2018)

\bibitem{sobhani2017dataset}
Sobhani, P., Inkpen, D., Zhu, X.: A dataset for multi-target stance detection.
  In: Proc. Conference of the European Chapter of the Association for
  Computational Linguistics (2017)

\bibitem{sobhani2019exploring}
Sobhani, P., Inkpen, D., Zhu, X.: Exploring deep neural networks for
  multitarget stance detection. Computational Intelligence  \textbf{35}(1),
  82--97 (2019)

\bibitem{sulis2016figurative}
Sulis, E., Far{\'\i}as, D.I.H., Rosso, P., Patti, V., Ruffo, G.: Figurative
  messages and affect in twitter: Differences between\# irony,\# sarcasm and\#
  not. Knowledge-Based Systems  \textbf{108},  132--143 (2016)

\bibitem{taule2017overview}
Taul{\'e}, M., et~al.: {Overview of the task on stance and gender detection in
  tweets on Catalan independence at IberEval 2017}. In: Proc. Workshop on
  Evaluation of Human Language Technologies for Iberian Languages (2017)

\bibitem{wang2019survey}
Wang, R., Zhou, D., Jiang, M., Si, J., Yang, Y.: A survey on opinion mining:
  from stance to product aspect. IEEE Access  (2019)

\bibitem{wei2018multi}
Wei, P., Lin, J., Mao, W.: Multi-target stance detection via a dynamic
  memory-augmented network. In: Proc. ACM SIGIR Conference (2018)

\bibitem{Wei2016pkudblabAS}
Wei, W., Zhang, X., Liu, X., Chen, W., Wang, T.: pkudblab at semeval-2016 task
  6 : A specific convolutional neural network system for effective stance
  detection. In: SemEval@NAACL-HLT (2016)

\bibitem{zarrella2016mitre}
Zarrella, G., Marsh, A.: Mitre at semeval-2016 task 6: Transfer learning for
  stance detection. arXiv preprint arXiv:1606.03784  (2016)

\bibitem{zubiaga2018detection}
Zubiaga, A., Aker, A., Bontcheva, K., Liakata, M., Procter, R.: Detection and
  resolution of rumours in social media: A survey. ACM Computing Surveys (CSUR)
   \textbf{51}(2), ~32 (2018)

\end{thebibliography}

\end{document}